\documentclass[preprint,11pt]{elsarticle}
\usepackage[margin=0.8in]{geometry}
\usepackage{amssymb}
\usepackage{amsmath}
\usepackage{array}
\usepackage{longtable}
\usepackage{subcaption}
\usepackage{hyperref}
\usepackage[dvipsnames]{xcolor}
\hypersetup{colorlinks=false,	linkbordercolor=white,	urlbordercolor=white,	pdfborder={0 0 0}}
\journal{}

\begin{document}

\begin{frontmatter}

\title{PsOCR: Benchmarking Large Multimodal Models for Optical Character Recognition in Low-resource Pashto Language}

\author[label1,label2]{Ijazul Haq\corref{cor1}}
\author[label1]{Yingjie Zhang}
\author[label3,label4]{Muhammad Saqib}

\cortext[cor1]{Corresponding author: Email address (hanjie@sjtu.edu.cn)}

\affiliation[label1]{
  organization={Shien-Ming Wu School of Intelligent Manufacturing, South China University of Technology},
  city={Guangzhou},
  postcode={511442},
  country={China}
}
\affiliation[label2]{
  organization={Artificial Intelligence Department, Guangdong CAS Angels Biotechnology Co. Ltd.},
  city={Foshan},
  country={China}
}
\affiliation[label3]{
  organization={Department of Software Engineering, Faculty of Electrical \& Computer Engineering,
  University of Engineering \& Technology},
  city={Peshawar},
  postcode={25000},
  country={Pakistan}
}
\affiliation[label4]{organization={Zirak AI}}
	
\begin{abstract}
This paper evaluates the performance of Large Multimodal Models (LMMs) on Optical Character Recognition (OCR) for the low-resource Pashto language. Pashto OCR is challenging due to its cursive Perso-Arabic script and the scarcity of large-scale annotated datasets. To address these challenges, we introduce PsOCR, a large-scale synthetic Pashto OCR dataset containing one million images annotated at the word, line, and document levels. PsOCR includes extensive variability across 1,000 font families, font sizes, colors, image resolutions, and layouts. A benchmark subset of 10,000 images is used to evaluate several state-of-the-art LMMs, including Llama, Florence, Qwen-3B/7B, GPT-4o, Gemini, Claude, and Grok, under zero-shot settings. Experimental results demonstrate that Gemini achieves the best overall performance, while Qwen-7B stands out among open-source models. This work provides valuable insights into the capabilities and limitations of current LMMs for Pashto OCR and establishes a foundation for future research in languages with similar scripts.
\end{abstract}

\end{frontmatter}

\section{Introduction}
\label{sec:introduction}
Optical Character Recognition (OCR) is essential for converting scanned and image-based documents into machine-readable text, underpinning digital archiving, automated indexing, and large-scale document analytics. Traditional OCR engines rely on hand-crafted rules and language-specific resources, while modern deep-learning approaches, using convolutional neural networks and transformer architectures, deliver remarkable performance. However, these methods depend heavily on extensive annotated corpora and lexicons, which are scarce or nonexistent for many languages \cite{ref1}. As a result, OCR performance degrades sharply for low-resource languages, highlighting the need for tailored datasets and models. In recent years, the research community has begun addressing these gaps: \cite{ref2} has worked on 60 low-resource languages; \cite{ref3} has focused on handwritten Tamil, Kurdish, Swahili, and Amharic scripts. Similarly, Ethiopian \cite{ref4}, Indic languages \cite{ref5}, \cite{ref6} and Khmer \cite{ref7} have also been explored.

Pashto is an Indo-European language of the Perso-Arabic script family, spoken by over 50 million people worldwide. It is the official language of Afghanistan, and the second-largest language of Pakistan by the number of native speakers \cite{ref8}. Written in cursive script from right to left (RTL), Pashto comprises 44 letters that can take up to four contextual forms: initial, medial, final, and isolated, which raises considerable challenges for OCR \cite{ref9}. Pashto also makes use of ligatures and diacritical marks, expanding its range of glyph shapes \cite{ref10}. Furthermore, there are no clear word boundaries in Pashto, and challenges arise from the inconsistent use of diacritical marks that further complicates the training of OCR models to recognize diverse orthographic conventions \cite{ref11}, \cite{ref12}, \cite{ref13}.

Synthetic data has emerged as an invaluable resource for addressing data scarcity issues in OCR research, particularly for low-resource languages and scripts. Generating synthetic datasets allows researchers to develop and benchmark robust AI models without the substantial time and resource investments required for manual annotation. This approach has proven effective across diverse languages and scripts; for instance, \cite{ref14} demonstrated that incorporating synthetic data alongside real-world data can lead to performance improvements. Similarly, \cite{ref15} introduced a comprehensive synthetic dataset for 23 Indic languages, significantly improving accuracy through fine-tuning on synthetic samples. Moreover, \cite{ref16} applied synthetic data generation techniques for Arabic OCR, facilitating extensive system evaluation and comparative studies. These studies underscore the importance and effectiveness of synthetic data in enhancing OCR model performance. In this work, we introduce PsOCR, a comprehensive dataset for training and evaluating LMMs on Pashto OCR tasks. PsOCR is composed of one million synthetic images, annotated at word, line, and document levels, featuring around a thousand unique font families, varied color schemes, and diverse layouts. A curated subset of 10K images is used as an evaluation benchmark to examine the performance of various flagship LMMs. We have evaluated a total of seven models, including four open-source models: Llama \cite{ref19}, Florence \cite{ref21}, and Qwen (3B and 7B) \cite{ref22}, and four proprietary models: Grok \cite{ref23}, Claude \cite{ref24}, GPT-4o \cite{ref25}, and Gemini \cite{ref26}. Detailed experimental analyses reveal each model’s strengths and weaknesses in Pashto text extraction, paving the way for further research in this domain. 

The key contributions of this study are as follows:
\begin{itemize}
	\item Introduced the first publicly available comprehensive Pashto OCR dataset consisting of one million synthetic images annotated at word, line, and document-level granularity, covering extensive variations including 1000 unique font families, diverse colors, image sizes, and text layouts.
	\item Developed the first publicly available OCR benchmark comprising 10K images, facilitating systematic evaluation and comparison of OCR systems for the low-resource Pashto.
	\item Conducted a pioneering evaluation and comparison of state-of-the-art LMMs on Pashto OCR, providing crucial insights into their zero-shot capabilities, strengths, and limitations for low-resource languages written in Perso-Arabic scripts.
\end{itemize}

\section{Related Work}
\label{related_work}
\subsection{Pashto OCR}
One of the foundational works in Pashto OCR is \cite{ref10}, identifying the primary recognizable units in Pashto script. They suggested “ligature” as the basic unit for OCR and claimed that 7,681 basic shapes were adequate for representing all Pashto ligatures, thus simplifying recognition challenges. \cite{ref27} proposed a robust OCR method addressing scale, rotation, and location invariances in Pashto script; they developed a holistic recognition framework and created a dataset of 8K images covering 1K unique ligatures. Recognizing dataset availability as a significant bottleneck, \cite{ref28} introduced the KPTI database containing 17K handwritten and printed text-line images, facilitating benchmarking of deep-learning models. \cite{ref29} used Sequential Minimal Optimization with horizontal and vertical projections for line, word, and character segmentation and Local Binary Patterns for feature extraction; this model achieved satisfactory performance and emphasized the importance of precise segmentation and robust feature extraction. \cite{ref30} leveraged transfer learning with pre-trained CNNs, highlighting the efficacy of fine-tuned deep-learning models augmented by data-enhancement techniques. \cite{ref31} developed a CNN-based classifier specifically targeting handwritten numerals, capable of handling variations in style and orientation. Similarly, \cite{ref32} and \cite{ref33} used CNNs and developed their own benchmark datasets for evaluation. \cite{ref34} tested feed-forward neural networks with varying ReLU layers on a custom dataset, providing crucial insights into neural architectures, while \cite{ref32} used LSTM for character recognition. \cite{ref35} developed the publicly accessible HPCID dataset, comprising 15K handwritten samples, offering critical resources for training and evaluation. Lastly, \cite{ref36} created the PHTI dataset, encompassing 36K segmented text-line images from diverse genres and writers, filling a substantial gap in available Pashto handwritten resources. A summary of isolated character datasets for Pashto OCR is given in Table~\ref{tab1}, while Table~\ref{tab2} compares our dataset with previous related works.

\begin{table*}[t]
\centering
\small
\caption{Summary of isolated characters datasets for Pashto OCR.}
\label{tab1}
\label{tab:isolated_datasets}
\setlength{\tabcolsep}{12pt}
\renewcommand{\arraystretch}{1.2} 
\begin{tabular}{l l l r l l l}\hline
\textbf{Ref.}  & \textbf{Type} & \textbf{Size} & \textbf{Classes} & \textbf{Granularity} & \textbf{Public} \\\hline
Ahmad et al. \cite{ref10}  & Synthetic & 8K & 1,000 ligatures & Ligature & Yes \\
FAST-NU \cite{ref30}  & Synthetic & 4K & 1,000 ligatures & Ligature & Yes \\
Ullah et al. \cite{ref29}  & Print & 5K & - & Character & No \\
Uddin et al. \cite{ref34}  & Handwritten & 43K & 43 characters & Character & Yes \\
Poha \cite{ref31} & Handwritten & 26.4K & 44 chars+10 digits & Character & Yes \\
Khan et al. \cite{ref32}  & Handwritten & 4.5K & 44 characters & Character & No \\
Rehman et al. \cite{ref33}  & Handwritten & 106K & 53 symbols & Character & Partial \\
Khan et al. \cite{ref67} & Handwritten & 50K & 10 digits & Digit & Yes \\\hline
\end{tabular}
\end{table*}

\begin{table*}[t]
\centering
\small
\caption{Comparison of text-level Pashto OCR datasets.}
\label{tab2}
\label{tab:text_ocr_datasets}
\setlength{\tabcolsep}{16pt}
\renewcommand{\arraystretch}{1.2} 
\begin{tabular}{l c c c}\hline
\textbf{Attribute} & \textbf{KPTI} \cite{ref28} & \textbf{PHTI} \cite{ref36} & \textbf{PsOCR (Ours)} \\\hline
Dataset Type & Real (Scanned) & Real (Handwritten) & Synthetic \\
Scale & 17K lines & 36K lines & 5.89M Lines/1M images\\
Font Color Variation & $\times$ & $\times$ & $\checkmark$ \\
Bg Color Variation & $\times$ & $\times$ & $\checkmark$ \\
Font Family Variation & $\times$ & $\times$ & $\checkmark$ \\
Font Size Variation & $\times$ & $\times$ & $\checkmark$ \\
Word-level Annotation & $\times$ & $\times$ & $\checkmark$ \\
Line-level Annotation & $\checkmark$ & $\checkmark$ & $\checkmark$ \\
Doc-level Annotation & $\times$ & $\times$ & $\checkmark$ \\
Writer Diversity & $\times$ & $\checkmark$ & $\times$ \\
Skew/Rotation & $\checkmark$ & $\checkmark$ & $\times$ \\\hline
\end{tabular}
\end{table*}

\subsection{Synthetic OCR Datasets}
Synthetic OCR datasets address annotated-data scarcity in low-resource languages. \cite{ref16} developed a method for generating synthetic Arabic OCR datasets, addressing unique script characteristics, facilitating comprehensive system testing and model comparisons. \cite{ref37} examined synthetic data methods for post-OCR correction, proposing a technique using computer vision-based glyph similarity algorithms. \cite{ref14} investigated synthetic data’s role in enhancing OCR model performance on the SROIE dataset; by combining synthetic data with real-world data, their model achieved approximately 32\% performance improvement. \cite{ref15} introduced a large-scale synthetic OCR benchmark dataset for 23 Indic languages, featuring varied fonts, sizes, colors, and backgrounds. Fine-tuning OCR models on their dataset improved accuracy by approximately 1\%. \cite{ref38} proposed RoundTripOCR, a synthetic data generation technique for post-OCR error correction in low-resource Devanagari languages, significantly enhancing OCR accuracy. \cite{ref39} and \cite{ref40} developed synthetic datasets for text localization in natural images.

\subsection{OCR Benchmarks for LMMs}
Under the large umbrella of OCR, various benchmarks are available for different tasks, such as for Visual Question Answering (VQA): TextOCR \cite{ref41}, TextCaps \cite{ref42}, ST-VQA \cite{ref43,ref17}, OCR-VQA \cite{ref44}, TextVQA \cite{ref45}, DocVQA \cite{ref46}, InfographicVQA \cite{ref47}, ChartQA \cite{ref48}, MTVQA \cite{ref49}, MultipanelVQA \cite{ref50}, EST-VQA \cite{ref51}; for rich-text image understanding: LLaVAR \cite{ref52}, MMR \cite{ref53}; and for reasoning: MM-GNN \cite{ref54,ref18}, textKVQA \cite{ref55}, OCRBench \cite{ref56}. Various other benchmarks are available for specific purposes, such as KITAB-Bench \cite{ref57} and CAMEL-Bench \cite{ref58} for Arabic OCR tasks, MOTBench \cite{ref59} for menu understanding and translation, CC-OCR \cite{ref60} for literacy, and Fox \cite{ref61,ref20} for fine-grained and multi-page document understanding. However, there is no significant prior work comparable to ours, which specifically focuses on benchmarking LMMs for text extraction in the low-resource Pashto language.

\section{Dataset Development}
\label{dataset_development}
\subsection{Text Corpus Collection}
Pashto has limited structured textual content available for large-scale NLP and OCR tasks. Developing an OCR dataset at a significant scale requires a substantial and high-quality raw text corpus. To address this requirement, we collected Pashto text from three primary sources. Firstly, we attempted extraction from the Common Crawl (CC) \footnote{\underline{http://www.ieee.org/authortools}} corpus. Although CC is extensive, the availability of Pashto text is extremely low; filtering a 15,000 GB dataset specifically for Pashto (ISO-639-3 language code: “pus”) yielded only \(\approx\)1GB of text, constituting about 0.008\% of the entire corpus. This indicates that relying solely on the CC corpus is inefficient for large-scale Pashto text collection. Secondly, we crawled open-source websites abundant in Pashto content, significantly supplementing our corpus. Third, we incorporated existing text resources from Twitter, books, and news websites used in previous studies \cite{ref11}, \cite{ref62}, \cite{ref63}, which constitute the major portion of our corpus. By combining these three data sources, we created a text corpus, sufficient for building a comprehensive OCR dataset.

\subsection{Text Cleaning and Preprocessing}
The text corpus underwent rigorous data cleaning and preprocessing steps to ensure its quality and usability. Initially, we removed extraneous data elements such as URLs, HTML tags, and longer chunks of foreign language text. Following this, normalization processes were applied to very large numerical values, repetitive line breaks, excessive spaces, emojis, and other special characters. Nevertheless, we intentionally maintained a controlled threshold of noise within the corpus rather than achieving absolute cleanliness. The rationale behind preserving minor noise was to enhance the versatility and robustness of OCR models trained on this dataset, enabling them to handle and generalize better during inference on real-world, imperfect text samples. Ultimately, the corpus was segmented into one million text chunks, where each chunk varied from as short as one sentence to as long as several paragraphs, facilitating diverse textual representation.

\subsection{Text to Image Conversion}
To convert textual content into images suitable for OCR model training, we adopted an automated method using Python scripts. Initially, each of the one million Pashto text chunks was programmatically converted into individual HTML pages. Subsequently, diverse yet controlled random styling, via Cascading Style Sheets (CSS), was applied to each HTML page utilizing Python and JavaScript scripts. This variability in style simulated realistic document formatting scenarios, significantly increasing dataset diversity. Finally, we utilized the Selenium\footnote{Selenium: \underline{https://pypi.org/project/selenium/ }} library to render these styled HTML pages. Each HTML page was captured as a PNG image screenshot, resulting in one million images of varying dimensions, aspect ratios, and visual styles, closely mimicking real-world document variability and complexities.

\subsection{Dataset Composition}
\subsubsection{Granularity}
\begin{figure*}[!t]
	\centering
	\includegraphics[width=\textwidth]{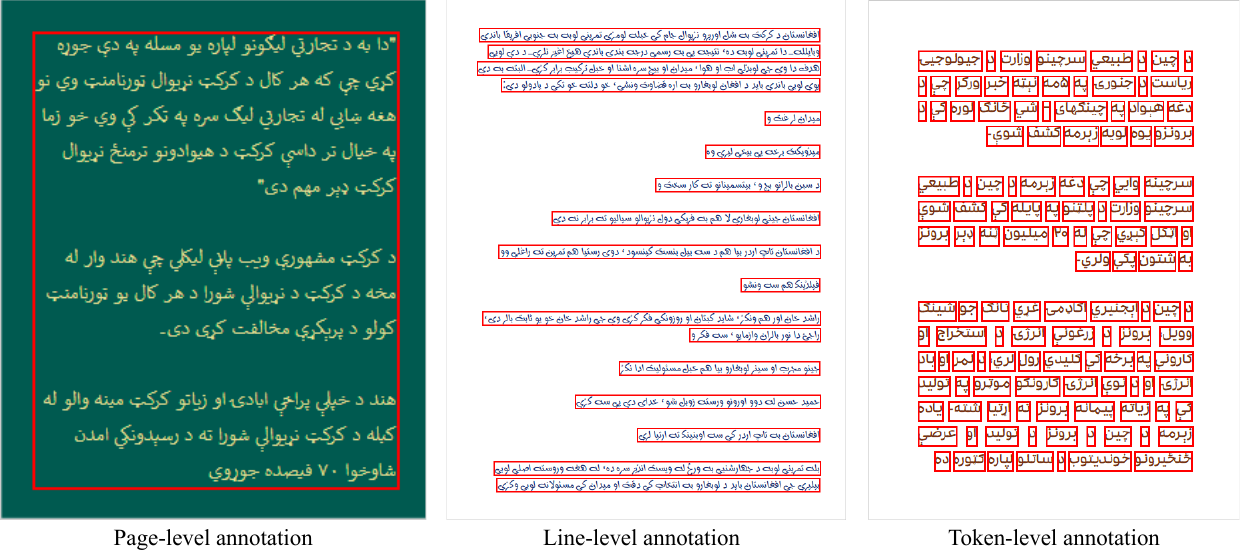}%
	\caption{Sample images from the dataset showing different levels of granularity and bounding box annotation}
	\label{fig1}
\end{figure*}

\begin{figure*}[!t]
	\centering
	\includegraphics[width=\textwidth]{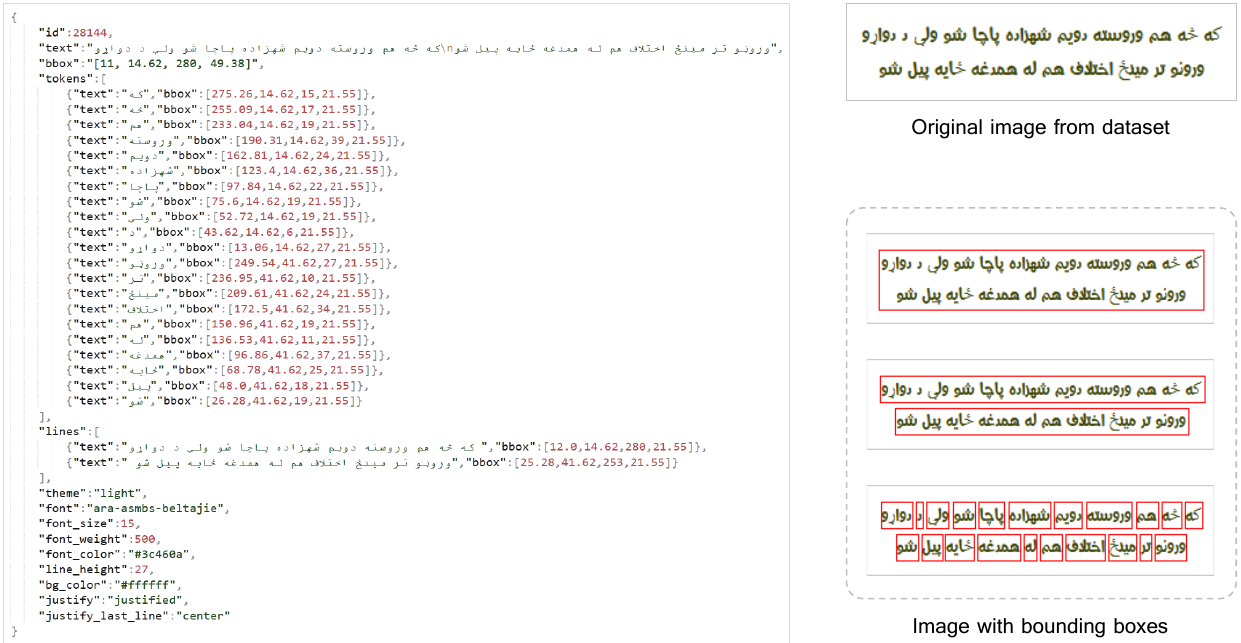}%
	\caption{A sample image from the dataset in PNG format, with the corresponding details and annotation information in JSON format}
	\label{fig2}
\end{figure*}

The PsOCR dataset was explicitly designed with the architectural diversity of AI models in mind, including both CNNs and Transformer-based architectures. The annotation information is provided at three levels of granularity: page-level, line-level, and token-level, as shown in Figure \ref{fig1}. Each annotation includes precise bounding boxes (bbox), characterized by four numerical attributes: (X, Y, width, height) as shown in Figure \ref{fig2}. Here, the coordinates (X, Y) represent the top-left corner of each bounding box, while width and height denote the respective horizontal and vertical dimensions measured in pixels. Page-level annotations comprise a single bounding box encapsulating all text present on an image page. Line-level annotations include individual bounding boxes for each distinct text line within a page. Similarly, token-level annotations define bounding boxes around every space-separated chunk of characters. This structured and rich annotation schema significantly enhances the dataset’s applicability across various OCR scenarios and training methodologies, supporting diverse granularity-focused tasks.

\subsection{Font Variation}
\begin{figure*}[t]
	\centering
	\includegraphics[width=\textwidth]{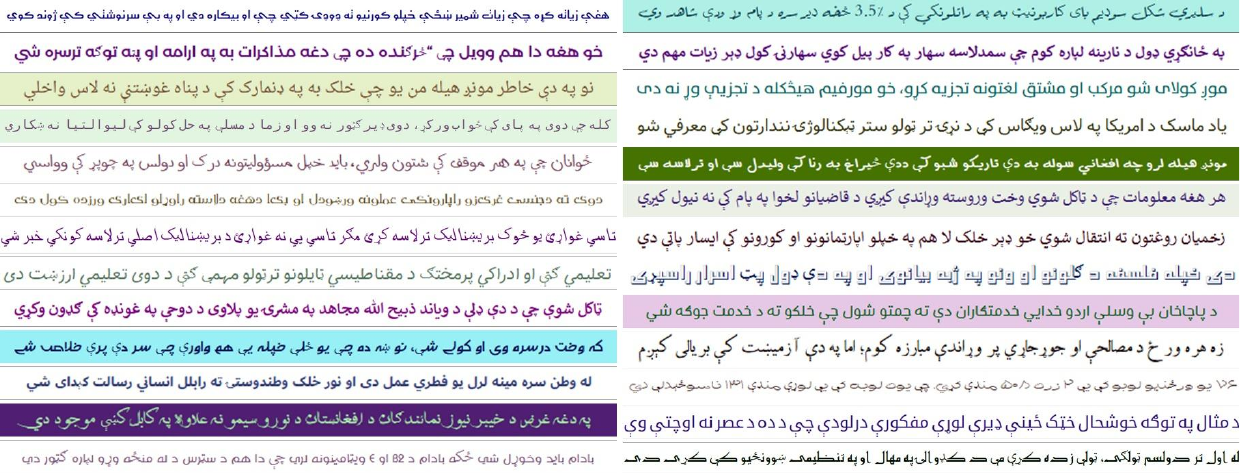}
	\caption{A few examples of font families in PsOCR dataset}
	\label{fig3}
\end{figure*}

Recognizing that font characteristics substantially influence OCR performance, we emphasized extensive font diversity within our dataset. Initially, approximately 3,000 Pashto-compatible font families were collected from publicly accessible sources. We carefully reviewed and filtered these fonts, removing any proprietary or non-freely distributable ones. Additionally, we manually inspected and removed fonts that were difficult to read. Duplicate fonts and those nearly identical in style and appearance were also removed to avoid redundancy. This rigorous selection process resulted in the inclusion of 1,000 distinct font families; some examples are shown in Figure \ref{fig3}. Furthermore, we ensured font sizes in the dataset ranged between 11px and 30px, providing suitable variability. We also varied font width by controlling text boldness through CSS numerical values ranging from 600 to 900. Such comprehensive font variation makes our PsOCR dataset a robust resource capable of training OCR models adept at handling extensive font-related variability encountered in practical OCR tasks.

\subsection{Images Size and Aspect Ratio}
Image size and aspect ratio are critical parameters for simulating realistic OCR scenarios and significantly influence model performance. To control variability, we predefined image widths via CSS, randomly selecting from 200, 300, 400, 500, 600, 700, and 800px, while heights were set to “auto,” resulting in variations based on text length, font size, line height, and the number of line breaks. The final rendered width and height include additional padding around the text area, slightly exceeding the original text element dimensions. Consequently, the images exhibit diverse aspect ratios. Figure \ref{fig5} (A) shows the distribution of image sizes, Figure \ref{fig5} (C) the histogram of aspect ratios, and Figure \ref{fig5} (B) the histogram of image file sizes, which affects storage requirements.

\subsection{Themes and Colors}
The PsOCR dataset encompasses a broad spectrum of color combinations, crucial for improving model adaptability to various visual environments. Primarily, two color themes were included: “Dark” and “Light”; in the “Dark” theme, a darker background is paired with lighter-colored text, and vice versa. Figure \ref{fig4} (A) depicts the thematic distribution across the dataset. The dataset contains a total of 232 dark and 271 light colors; the top 200 most frequent dark and light colors are shown in Figure \ref{fig4} (B) and (C), respectively. Given these color counts, theoretically, approximately 126K unique color combinations could have been generated (2 themes × 232 dark × 271 light). However, the actual number of unique combinations included is around 66K. This discrepancy arises from our controlled color selection strategy designed to ensure optimal readability and visual contrast. Specifically, the following procedure was used for selecting colors for each image:

\begin{figure*}
	\centering
	\includegraphics[width=\textwidth]{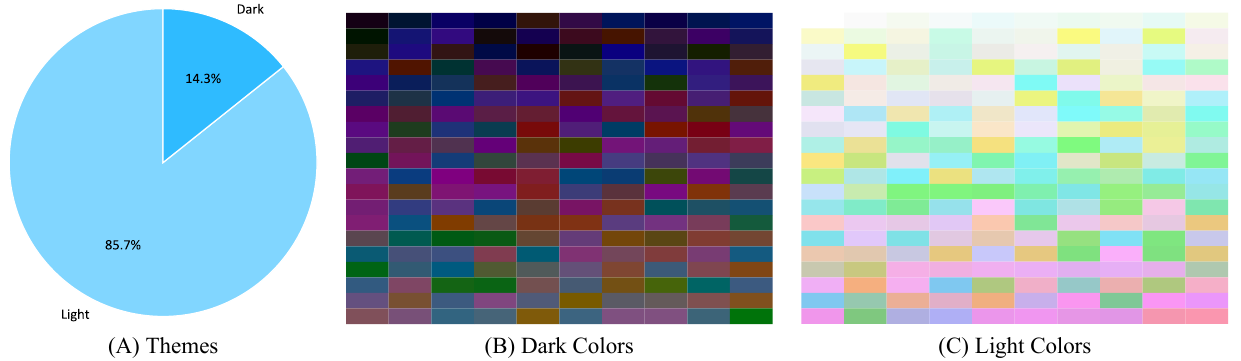}
	\caption{Colors and themes in the dataset and their ratio}
	\label{fig4}
\end{figure*}

\begin{itemize}
	\item\textbf{Step 1:} Randomly select a theme for each image. The probability of selecting the “Light” theme was set six times higher than selecting the “Dark” theme, reflecting common readability practices in real-world documents.
	\item\textbf{Step 2:} If the selected theme is Dark, randomly choose a dark color for the background and a light color for the text; if the theme is Light, choose a light background and a dark text color.
	\item\textbf{Step 3:} Calculate the luminance for both selected background and font colors. The luminance ratio is computed by dividing the higher luminance value by the lower luminance value of the color pair.
	\item\textbf{Step 4.} If the luminance ratio is less than 6, discard the selected color pair and return to Step 2. If the luminance ratio is \(\geq\) 6, accept the chosen background and font colors. This contrast threshold (\(\geq\) 6) was chosen empirically based on human readability perception across numerous samples.
\end{itemize}

This method ensures that all included color combinations provide clear visual contrast, facilitating effective recognition by OCR models.

\subsection{Other Variations}
In addition to the previously described attributes, several other variations were introduced, briefly explained as follows:
\begin{itemize}
	\item\textbf{Padding:} Padding denotes the empty space between the text content and the image borders. We allowed padding values of 10, 20, 30, 40, and 50px for all the images, while padding was limited to 10, 20, or 30px for images narrower than 200px to avoid excessive white space.
	\item\textbf{Text Alignment:} To simulate common document layouts, we randomly applied one of three text alignment options to each document: “right,” “justified,” or “center.” For images using justified alignment, we further varied the final line, aligning it either to the “right” or “center.”
	\item\textbf{Number of Lines and Paragraphs:} The dataset reflects organic variability in text structure by allowing each image to contain an arbitrary number of paragraphs and lines. Paragraph counts follow the natural segmentation of the source text, without artificial splitting or merging.
	\item\textbf{Line Height:} The vertical spacing between consecutive text lines was also varied across the dataset to mirror real-world document formatting.
\end{itemize}

\begin{figure*}
	\centering
	\includegraphics[width=\textwidth]{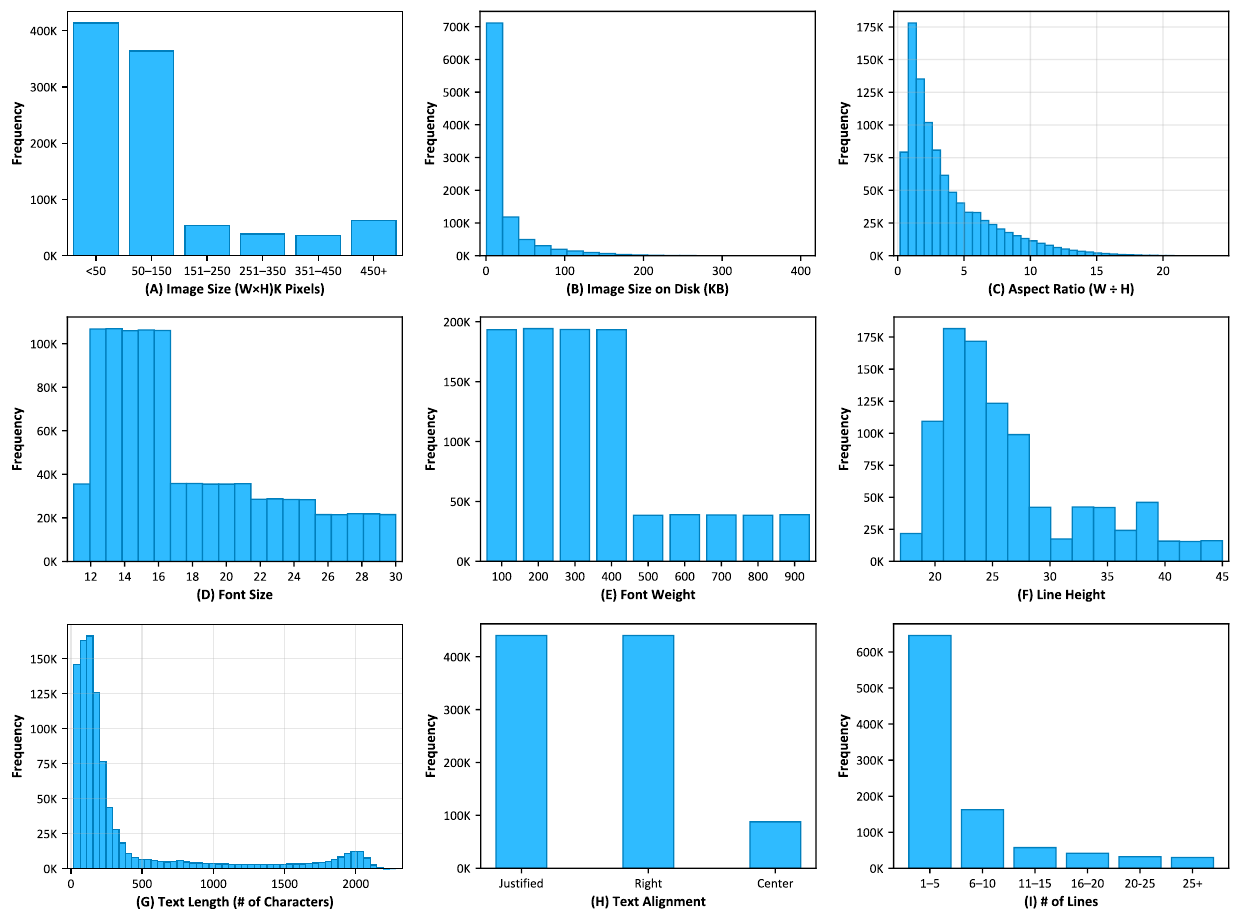}
	\caption{Dataset Statistics}
	\label{fig5}
\end{figure*}

\section{Validation and Evaluation}
\label{evaluation}
\subsection{Models Selection}
This is a pioneering study of LMMs evaluation on any Pashto benchmark, thus limited previous literature exists to guide model selection. Given the rapid growth in the number of available LMMs, careful selection criteria were essential to determine the most relevant models to include in our analysis. We primarily selected models based on their popularity in existing research literature, their documented performance on public leaderboards, and, for open-source models, their download frequency and current trends on platforms such as ModelScope\footnote{ModelScope: \underline{https://www.modelscope.cn}} and HuggingFace\footnote{HuggingFace: \underline{https://huggingface.co}}. Additionally, we ensured all selected open-source variants are of similar size in terms of parameter count; thus, we chose the 7B variants or, if unavailable, those closest to 7B.

In total, this study evaluates seven VLMs: four open-source (Llama, Florence, Qwen 3B, and Qwen 7B) and four proprietary (GPT-4o, Gemini, Claude, and Grok). A concise overview of these models, along with their parent organizations, specific variants, and parameter counts, is summarized in Table \ref{tab3}.

\begin{table*}
	\caption{Summary of the Evaluated Models}
	\label{tab3}
	\setlength{\tabcolsep}{20pt}
	\renewcommand{\arraystretch}{1.2} 
	\centering
	\begin{tabular}{l l l l}\hline
		Model & Org & Exact Variant & \# of Parameters \\\hline
		Llama \cite{ref19}    & Meta         & Llama-3.2-11B-Vision-Instruct & 10.7B \\ 
		Florence \cite{ref21} & Microsoft    & Florence-2-large             & 0.77B  \\ 
		Qwen-3B \cite{ref22}  & Alibaba      & Qwen2.5-VL-3B-Instruct       & 3.75B  \\ 
		Qwen-7B \cite{ref22}  & Alibaba      & Qwen2.5-VL-7B-Instruct       & 8.29B  \\\hline
		Grok \cite{ref23}     & X-AI         & grok-2-vision-1212           & –      \\ 
		Claude \cite{ref24}   & Anthropic    & claude-3-7-sonnet-20250219   & –      \\ 
		GPT-4o \cite{ref25}   & OpenAI       & GPT-4o (2024-08-06)          & –      \\ 
		Gemini \cite{ref26}   & Google       & gemini-2.0-flash             & –      \\\hline
	\end{tabular}
\end{table*}

\subsection{Experimental Setup}
The core objective of this study is to assess the zero-shot OCR performance of selected LMMs on our newly developed PsOCR benchmark. Neither pre-training nor fine-tuning was applied; models were evaluated directly in their original form. The inference pipeline for proprietary models: GPT-4o, Gemini, Claude, and Grok was developed through their respective APIs. Conversely, selected open-source models, including Llama, Florence, and Qwen, were downloaded and tested locally. Decisions regarding the use of APIs versus local inference were driven by model accessibility and available computational resources. For the experiment, we used Python-based tools and libraries such as openai\footnote{OpenAI Python API library: \underline{https://github.com/openai/openai-python}}, HuggingFace’s Transformers\footnote{HuggingFace Transformers: \underline{https://github.com/huggingface/transformers}}, Google’s GenAI SDK\footnote{Google Gen AI SDK: \underline{https://github.com/googleapis/python-genai }}, and Anthropic’s API\footnote{Anthropic API: \underline{https://docs.anthropic.com/en/api/getting-started}}. Local inference ran on a PC with an Intel Core i7 processor, 32GB of RAM, and an NVIDIA RTX 4080 GPU. During evaluation, each of the 10K images was provided one by one to each model along with a specifically engineered prompt, and the outputs were recorded. Although most evaluated models support conversational context, our experiment explicitly avoided using histories; the evaluation strictly adhered to a zero-shot setting, providing no contextual examples or prior information, to ensure an unbiased assessment of their innate OCR capabilities.

\subsection{Prompt Engineering}
\begin{figure}
	\centering
	\includegraphics[width=\columnwidth]{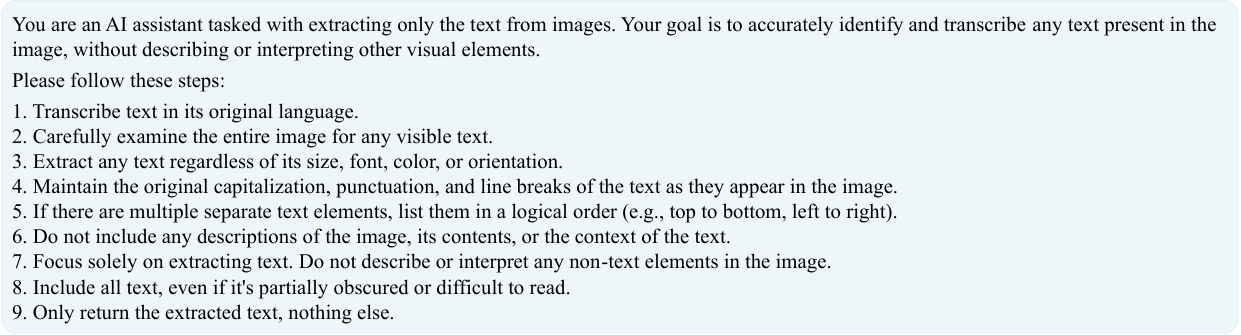}
	\caption{Prompt for Llama and Claude}
	\label{fig6}
\end{figure}

\begin{figure}
	\centering
	\includegraphics[width=\columnwidth]{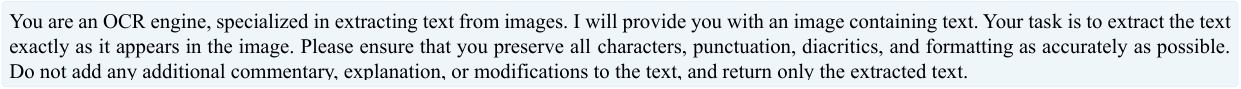}
	\caption{Prompt for GPT4-o, Gemini, Grok, Florence and Qwen}
	\label{fig7}
\end{figure}

A critical factor affecting evaluation results in zero-shot scenarios is the quality and specificity of the instructions provided to the models, known as “prompt.” Effective prompt design is essential to achieving accurate model responses. However, no universal standards or guidelines exist for crafting optimal prompts. Different models have been trained and fine-tuned by diverse research teams using varying methodologies, leading to notable differences in how models interpret and respond to prompts. Consequently, a prompt effective for one model may not be equally suitable for another. To address this, we invested considerable effort in designing customized prompts for each model. First we wrote several prompts following some standard guidlines, such as those mentioned in \cite{ref64,ref65}. Then we chose the prompt giving the best performance. After selecting the best prompts we adopted a trial-and-error approach to iteratively refine instructions to further maximize performance. A primary difficulty encountered during prompt engineering involved clearly instructing the models to return only the exact text displayed within the images, without providing additional contextual information, explanations, or translations. Several models demonstrated a tendency to translate the extracted text or include extraneous details. Thus, significant prompt refinement was necessary to explicitly discourage translation and ensure that the outputs precisely reflected the original Pashto text. Figure \ref{fig6} and Figure \ref{fig7} present two prompt examples that yielded the best results in our experiments.

\subsection{Post Processing Models' Responses}
Despite rigorous efforts in prompt engineering, some models still produced responses that did not strictly adhere to the desired output format, occasionally including additional comments or irrelevant contextual information. Therefore, after collecting predictions from all models, a manual verification step was performed to identify and correct invalid or improperly formatted responses. During this validation, it was observed that most models reliably provided responses in the expected format, except GPT-4o and the Llama model. The GPT-4o model, in particular, occasionally generated erroneous or incomplete responses due to content flagged as inappropriate by its internal safety filters. In the case of GPT-4o, such erroneous responses accounted for approximately 4\% of the total benchmark. Similarly, the Llama model demonstrated considerable difficulty in following the instructions, consistently producing responses contaminated with additional commentary. To handle these outputs, a semi-automatic cleaning procedure was employed to remove unnecessary text, isolating the predicted OCR content for accurate comparison with the ground-truth texts.

\subsection{Evaluation Metrics}
To comprehensively evaluate the document-level OCR performance of LMMs, we adopt two complementary categories of evaluation metrics: 
\textit{(1) transcription accuracy metrics} and \textit{(2) text similarity metrics}. 
This dual evaluation strategy allows us to assess both the exactness of OCR transcription and the high-level similarity between ground-truth and predicted texts.

\subsubsection{Transcription Accuracy Metrics}
Transcription accuracy metrics are standard in OCR research and measure how precisely the recognized text matches the ground-truth transcription at the character and word levels. These metrics are based on edit distance and explicitly penalize substitution, insertion, and deletion errors, making them well-suited for evaluating OCR performance in low-resource and cursive scripts such as Pashto. We employ the following two widely used transcription accuracy metrics.

\paragraph{Character Error Rate (CER)}
Character Error Rate measures the normalized edit distance between the predicted text and the ground-truth text at the character level. It is defined as:
\begin{equation}
\mathrm{CER} = \frac{S + D + I}{N},
\end{equation}
where $S$, $D$, and $I$ denote the number of character substitutions, deletions, and insertions, respectively, and $N$ is the total number of characters in the ground-truth text. Lower CER values indicate better OCR performance.

\paragraph{Word Error Rate (WER)}
Word Error Rate evaluates transcription errors at the word level and is computed analogously to CER:
\begin{equation}
\mathrm{WER} = \frac{S_w + D_w + I_w}{N_w},
\end{equation}
where $S_w$, $D_w$, and $I_w$ represent word-level substitutions, deletions, and insertions, and $N_w$ is the total number of words in the ground-truth document. WER reflects word-level usability of OCR output for downstream NLP applications.

\subsubsection{Text Similarity Metrics}
In addition to transcription accuracy metrics, we report text similarity metrics to analyze the overall similarity between predicted and ground-truth documents. Since this study evaluates transformer-based generative models, such metrics provide complementary insights into the models’ ability to generate text sequences that are globally similar to the reference, even when minor transcription errors are present. These metrics are reported as auxiliary measures and are not treated as direct indicators of OCR accuracy.

\paragraph{BLEU}
BLEU measures the precision of overlapping $n$-grams between the predicted text and the ground-truth text. It is computed as:
\begin{equation}
\mathrm{BLEU} = \mathrm{BP} \cdot \exp \left( \sum_{n=1}^{N} w_n \log p_n \right),
\end{equation}
where $p_n$ denotes the modified $n$-gram precision, $w_n$ are weighting factors, and $\mathrm{BP}$ is a brevity penalty that penalizes overly short predictions.

\paragraph{METEOR}
METEOR evaluates text similarity based on unigram alignment between the predicted and reference texts, incorporating both precision and recall. It is defined as:
\begin{equation}
\mathrm{METEOR} = F_{\text{mean}} \cdot (1 - P),
\end{equation}
where $F_{\text{mean}}$ is the harmonic mean of unigram precision and recall, and $P$ is a penalty term that accounts for fragmented alignments. 

\paragraph{Bag-of-Words (BoW) Similarity}
BoW evaluation measures text similarity by comparing the unordered sets of words in the predicted and ground-truth texts, ignoring word order. Each document is represented as a word frequency vector, and similarity is computed using cosine similarity:
\begin{equation}
\mathrm{BoW} = \frac{\mathbf{v}_{\text{gt}} \cdot \mathbf{v}_{\text{pred}}}
{\|\mathbf{v}_{\text{gt}}\| \, \|\mathbf{v}_{\text{pred}}\|},
\end{equation}
where $\mathbf{v}_{\text{gt}}$ and $\mathbf{v}_{\text{pred}}$ denote the BoW vectors of the ground-truth and predicted texts, respectively. 

\section{Results and Discussion}
\subsection{Models Performance Comparison}
\begin{figure*}[t]
	\centering
	\includegraphics[width=\textwidth]{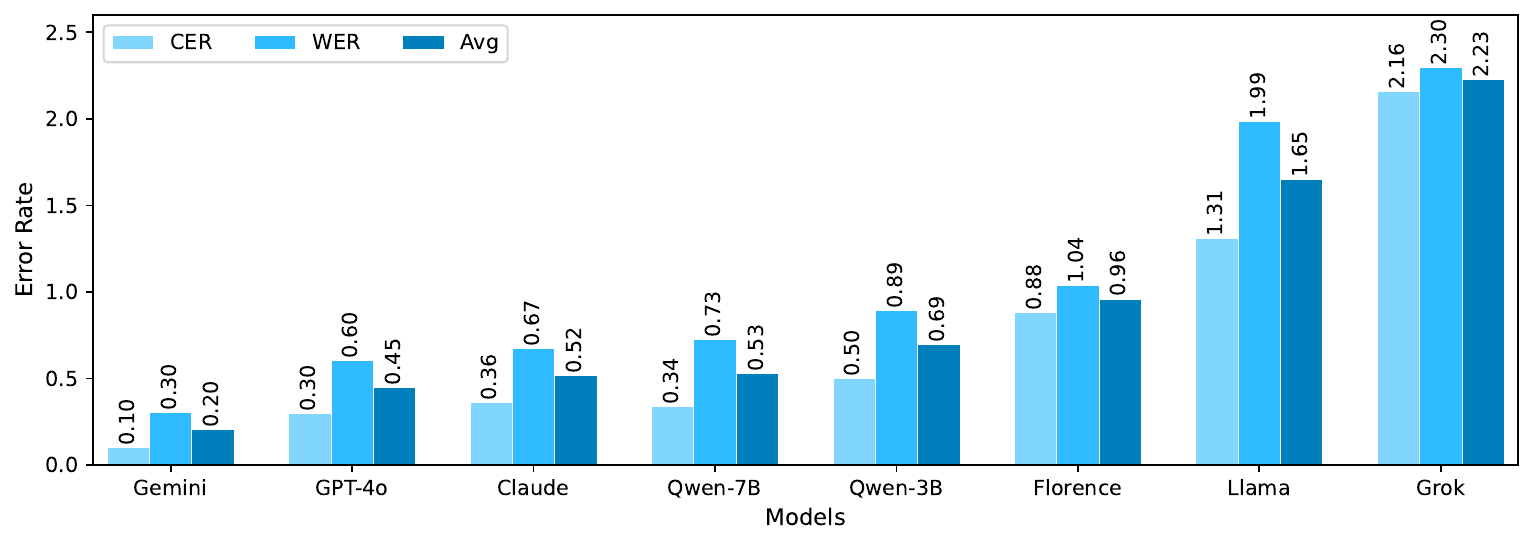}
	\caption{Visualization of CER and WER performance of LMMs on the Pashto OCR benchmark (lower is better).}
	\label{fig8}
\end{figure*}

Tables~\ref{tab4} and~\ref{tab5}, together with the visualization in Figure~\ref{fig8}, present the performance of the evaluated LMMs on the PsOCR benchmark. To provide a comprehensive assessment, we report results using two complementary categories of evaluation metrics: transcription accuracy metrics, namely CER and WER, and text similarity metrics, including BLEU, METEOR, and BoW similarity. This dual evaluation framework enables analysis of both exact transcription correctness and overall textual similarity between predicted outputs and ground-truth documents.

The transcription accuracy results in Table~\ref{tab4} reveal substantial performance differences among the models. Gemini achieves the best overall performance, with a CER of 0.10 and a WER of 0.31, indicating strong character- and word-level transcription accuracy in a zero-shot setting. Among the proprietary models, GPT-4o and Claude also demonstrate competitive performance. Within the open-source group, Qwen-7B emerges as the strongest performer, achieving a CER of 0.34 and a WER of 0.73, closely matching the performance of some proprietary systems. In contrast, models such as Llama and Grok exhibit high error rates, with CER and WER values exceeding 1.0. Across all models, WER is consistently higher than CER, suggesting that while character recognition is relatively robust, maintaining correct word boundaries and producing accurate word sequences remains a significant challenge for Pashto OCR.

The text similarity results reported in Table~\ref{tab5} largely align with the transcription accuracy trends, while offering complementary insights into global textual similarity. Gemini again outperforms all other models, achieving the highest scores, with an average similarity of 0.63. GPT-4o and Claude follow, with GPT-4o showing stronger overall similarity to the ground truth, particularly in terms of lexical overlap. Among open-source models, Qwen-7B consistently attains the highest similarity scores, reinforcing its position as a strong open-source baseline. It is important to note that these similarity metrics provide auxiliary evidence of how closely the generated text resembles the ground truth at a sequence and lexical level.

These results highlight several key observations. First, proprietary LMMs generally outperform open-source alternatives; however, Qwen-7B significantly narrows this gap and, in some cases, approaches the performance of closed-source models. Second, the persistent disparity between CER and WER across models underscores the difficulty of word-level OCR for Pashto, a language characterized by cursive writing, ligatures, and ambiguous word boundaries. Third, the strong correspondence between low transcription error rates and high text similarity scores suggests that models producing accurate transcriptions also tend to generate text that is globally coherent and lexically aligned with the ground truth. Collectively, these findings demonstrate that Gemini offers the strongest zero-shot OCR performance for Pashto, while Qwen-7B represents a promising and accessible open-source alternative, establishing solid baselines for future research on OCR in low-resource, Perso-Arabic script languages.

\begin{table*}[t]
	\caption{Comparative evaluation of LMMs on the Pashto OCR benchmark using CER and WER (lower is better).}
	\label{tab4}
    \setlength{\tabcolsep}{48pt}
	\renewcommand{\arraystretch}{1.2}
	\centering
	\begin{tabular}{l c c c}\hline
		Model & CER & WER & Avg \\\hline
		Llama      & 1.31 & 1.99 & 1.65 \\
		Florence   & 0.88 & 1.04 & 0.96 \\
		Qwen-3B    & 0.50 & 0.89 & 0.70 \\
		Qwen-7B    & 0.34 & 0.73 & 0.53 \\\hline
		Grok       & 2.16 & 2.30 & 2.23 \\
		Claude     & 0.36 & 0.67 & 0.52 \\
		GPT-4o     & 0.30 & 0.60 & 0.45 \\
		Gemini     & 0.10 & 0.31 & 0.20 \\\hline
	\end{tabular}
\end{table*}

\begin{table*}[t]
	\caption{Comparative evaluation of LMMs on the Pashto OCR benchmark using BLEU, METEOR, and BoW similarity metrics (higher is better).}
	\label{tab5}
	\setlength{\tabcolsep}{32pt}
	\renewcommand{\arraystretch}{1.2}
	\centering
	\begin{tabular}{l c c c c}\hline
		Model & BLEU & METEOR & BoW & Avg \\\hline
		Llama      & 0.07 & 0.23 & 0.27 & 0.19 \\
		Florence   & 0.00 & 0.00 & 0.00 & 0.00 \\
		Qwen-3B    & 0.08 & 0.22 & 0.25 & 0.18 \\
		Qwen-7B    & 0.12 & 0.32 & 0.34 & 0.26 \\\hline
		Grok       & 0.00 & 0.01 & 0.02 & 0.01 \\
		Claude     & 0.19 & 0.36 & 0.37 & 0.31 \\
		GPT-4o     & 0.28 & 0.49 & 0.49 & 0.42 \\
		Gemini     & 0.50 & 0.72 & 0.67 & 0.63 \\\hline
	\end{tabular}
\end{table*}

\begin{figure*}[t]
	\centering
	\includegraphics[width=\textwidth]{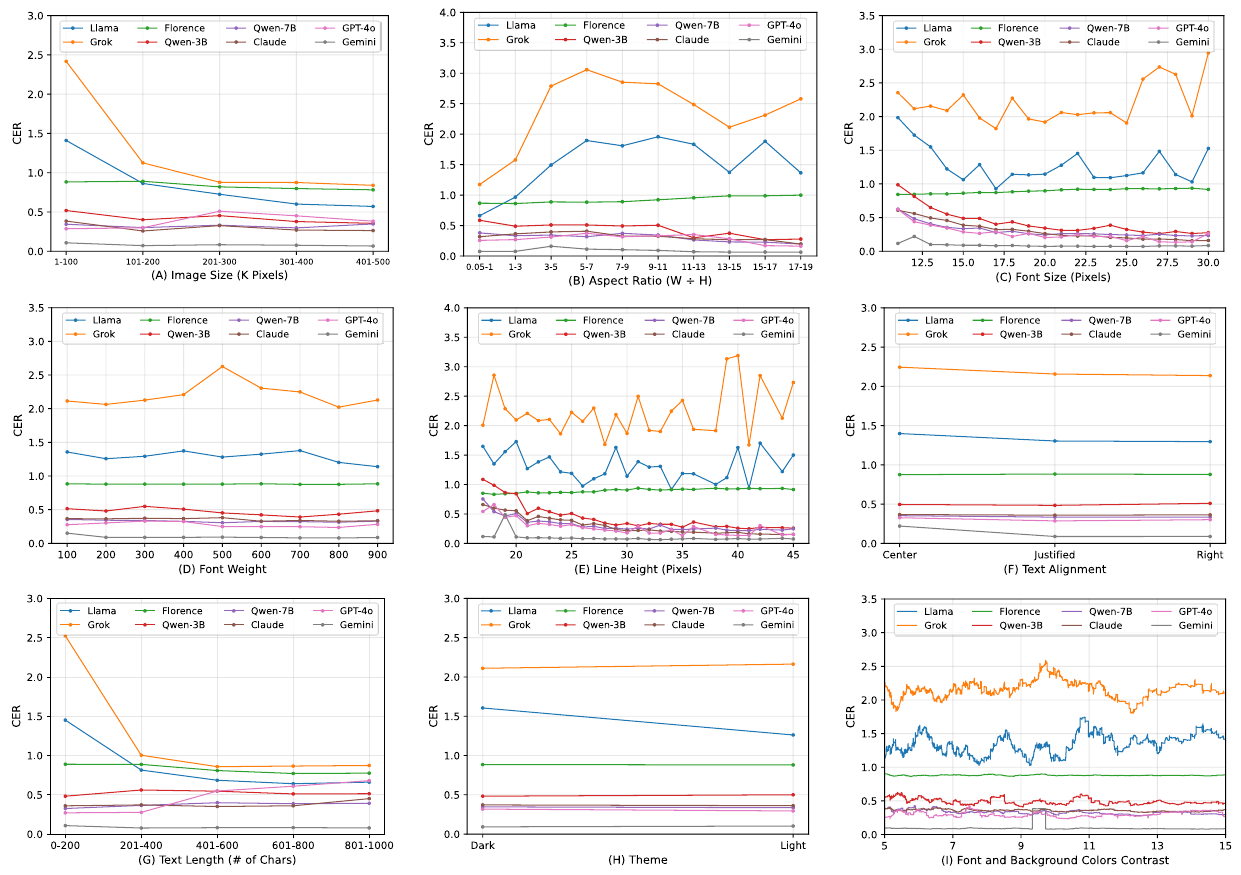}
	\caption{Effect of different variations on models' performance}
	\label{fig9}
\end{figure*}

\begin{figure*}
	\centering
	\begin{minipage}{0.49\textwidth}
		\centering
		\includegraphics[width=\linewidth]{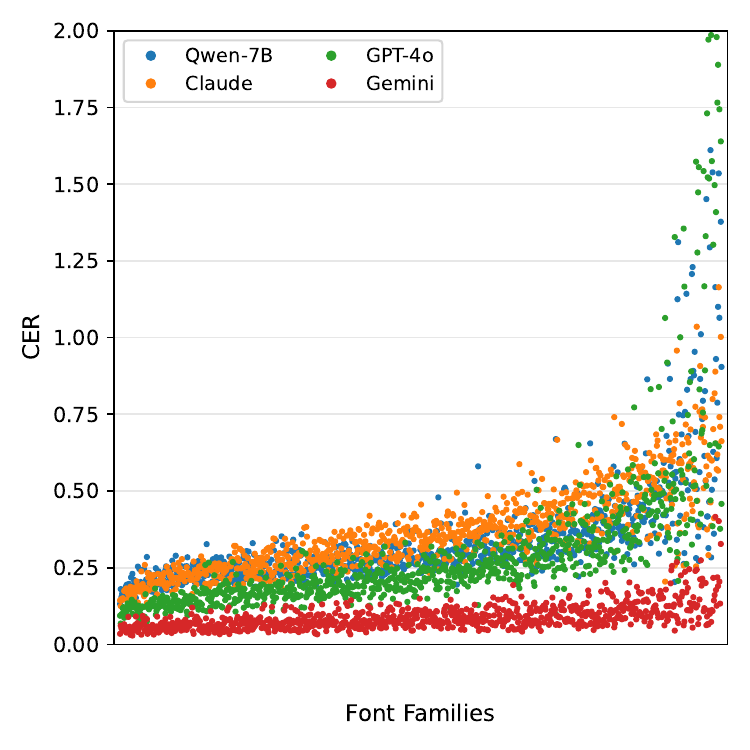}
		\caption{Performance variation across fonts}
		\label{fig10}
	\end{minipage}
	\hfill
	\begin{minipage}{0.49\textwidth}
		\centering
		\includegraphics[width=\linewidth]{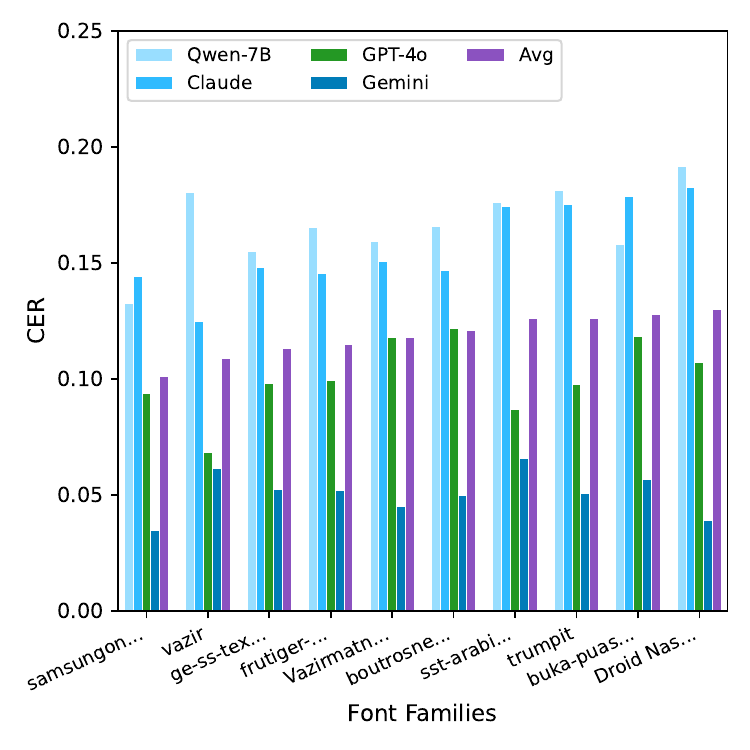}
		\caption{LMMs scores on top-10 best-performing fonts}
		\label{fig11}
	\end{minipage}
\end{figure*}

\begin{table*}
	\caption{List of top-10 best-performing font families\label{tab6}}
	\centering
	\setlength{\tabcolsep}{15pt}
	\renewcommand{\arraystretch}{1.1} 
\begin{tabular}{c l c c }
	\hline
	Rank & Font Name & Avg. CER & Font Style \\ \hline
	1.  & samsungonearabic-400              & 0.101 & \includegraphics[width=6cm]{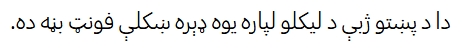}  \\
	2.  & vazir                             & 0.109 & \includegraphics[width=6cm]{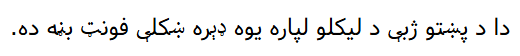}  \\
	3.  & ge-ss-text-medium                 & 0.113 & \includegraphics[width=6cm]{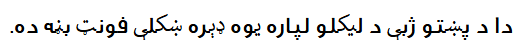}  \\
	4.  & frutiger-lt-arabic-45-light       & 0.115 & \includegraphics[width=6cm]{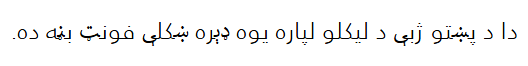}  \\
	5.  & Vazirmatn-Regular                 & 0.118 & \includegraphics[width=6cm]{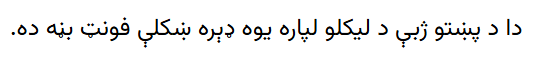}  \\
	6.  & boutrosnewsh1-bold                & 0.121 & \includegraphics[width=6cm]{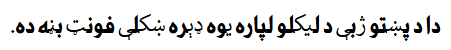}  \\
	7.  & sst-arabic-medium                 & 0.126 & \includegraphics[width=6cm]{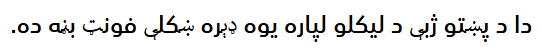}  \\
	8.  & trumpit                           & 0.126 & \includegraphics[width=6cm]{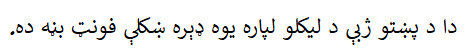}  \\
	9.  & buka-puasa-bersama-normal         & 0.128 & \includegraphics[width=6cm]{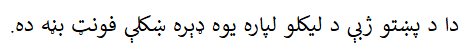}  \\
	10. & Droid Naskh-Regular               & 0.130 & \includegraphics[width=6cm]{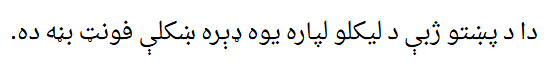} \\ \hline
\end{tabular}
\end{table*}

\begin{figure}[t]
	\centering
	\includegraphics[width=\columnwidth]{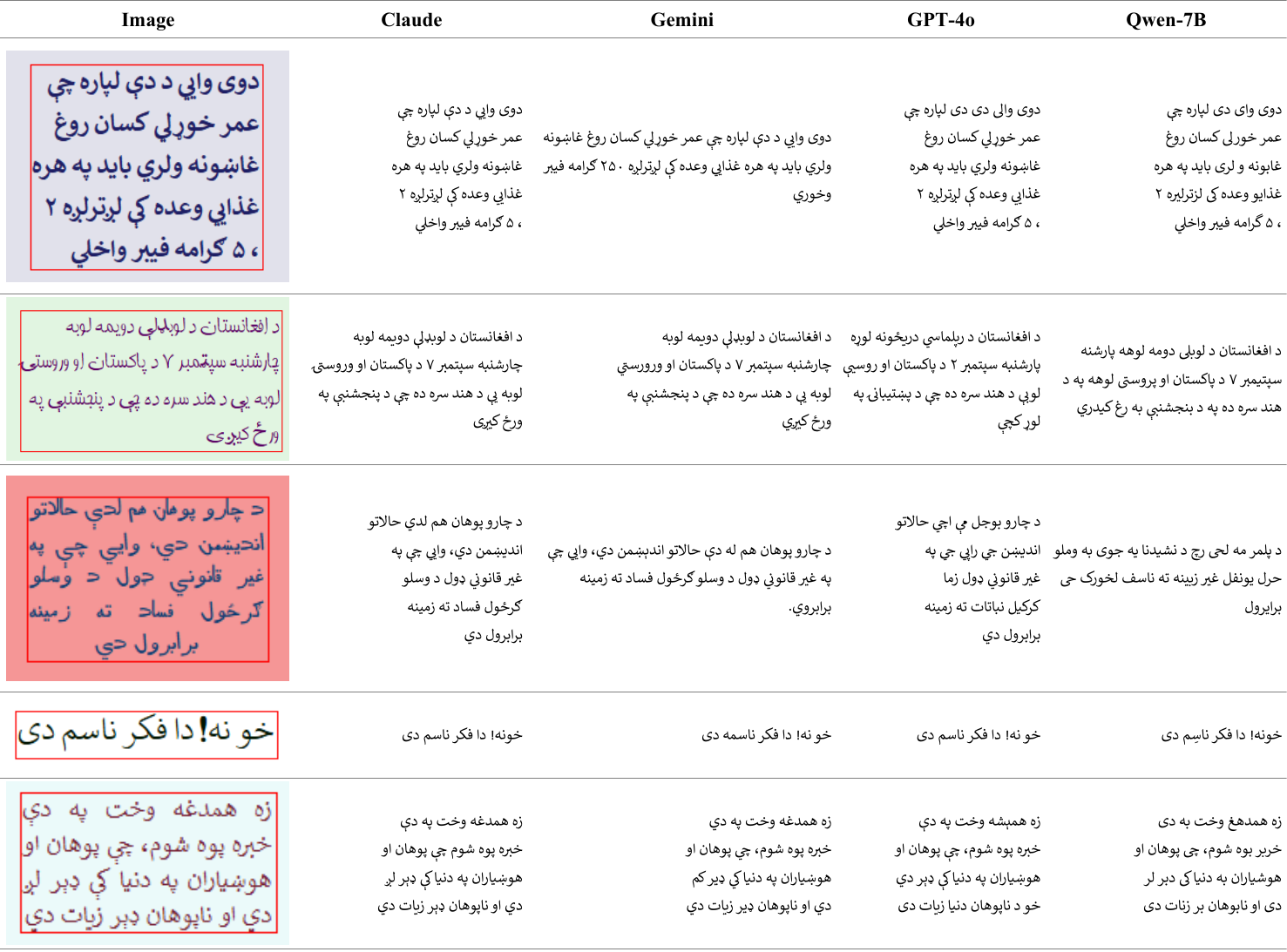}
	\caption{Example outputs of various models.}
	\label{fig12}
\end{figure}

\subsection{In-depth Results Analysis:}
To gain deeper insights into how various image and text properties influence Pashto OCR performance, we plotted each model’s average score against key image attributes, as shown in Figure \ref{fig9}. Figure \ref{fig12} lists some example outputs of various models. The following discussion summarizes the effect of each attribute on models’ performance.

\textbf{Effect of Image File Size:}
In Figure \ref{fig9} (A), we observe a modest upward trend in model performance as image size increases. While the overall trend is subtle, models such as Claude, Gemini, and Qwen-7B exhibit noticeable performance gains on the largest images (600-700K px).

\textbf{Effect of Image Aspect Ratio:}
Figure \ref{fig9} (B) shows that aspect ratio exerts minimal influence on OCR accuracy. Nevertheless, all the models demonstrate slightly better performance on wider images compared to taller ones.

\textbf{Effect of Font Size:}
Unlike image size and aspect ratio, font size clearly impacts accuracy. All models perform better on images with larger font sizes, with Qwen-3B showing the greatest sensitivity to this factor. Gemini, by contrast, maintains consistently high performance across the font sizes.

\textbf{Effect of Font Weight:}
As illustrated in Figure \ref{fig9} (D), variations in font weight (from light to bold) have negligible impact on model performance. The largely horizontal lines in this plot demonstrate that both thin and heavy typefaces are handled similarly by the models.

\textbf{Effect of Line Height:}
Line spacing proved to be one of the most influential factors. Models such as Qwen, Claude, and GPT-4o struggle completely on images with very tight line spacing (\(\leq\)20px), yielding average scores near zero. Performance improves steadily as line spacing increases.

\textbf{Effect of Text Alignment:}
Text alignment shows little overall effect on model accuracy. A slight improvement is visible for “justified” alignment and a minor performance dip for “left” alignment, which aligns with Pashto’s RTL writing direction.

\textbf{Effect of Text Length:}
As Figure \ref{fig9} (G) demonstrates, text length has minimal impact on most models’ performance. An exception is GPT-4o, whose average score declines significantly on images containing longer passages, suggesting that this model may struggle to maintain accuracy for longer text sequences.

\textbf{Effect of Theme and Color Contrast:}
Figure \ref{fig9} (H) and (I) reveal that neither overall theme nor specific foreground-background color pairings significantly affect OCR performance.

\textbf{Effect of Font Family:}
Font family exerts the strongest influence on OCR accuracy, as shown in Figure \ref{fig10}. Given the dataset’s diverse font families, models exhibit wide performance variation. These findings highlight font diversity as one of the primary challenges in Pashto OCR. Figure \ref{fig11} presents a comparative analysis of models’ performance on the top 10 fonts, and Table \ref{tab6} lists those fonts.

\section{Conclusion and Future Work}
In this work, we presented the first large-scale evaluation of state-of-the-art LMMs on a newly developed Pashto OCR benchmark. Our dataset, comprising one million synthetically generated images annotated at the word, line, and document levels, represents the first publicly available resource of its kind for systematic OCR evaluation in the low-resource Pashto language. We evaluated four open-source models (Llama, Florence, Qwen-3B, and Qwen-7B) alongside four proprietary models (GPT-4o, Gemini, Claude, and Grok) under zero-shot settings using CER, WER, BLEU, METEOR, and BoW similarity metrics. Experimental results show that Gemini consistently achieves the best overall performance, attaining the lowest CER of 0.10 and WER of 0.31 and the highest text similarity scores, while among open-source models, Qwen-7B stands out as the strongest performer. These findings underscore both the growing zero-shot OCR capabilities of current LMMs for cursive, ligature-rich scripts such as Pashto and the promise of open-source models like Qwen-7B as strong foundations for future fine-tuning and adaptation.

This study also has some known limitations. First, the dataset consists solely of computer-generated text and does not include handwritten samples. Additionally, the image backgrounds are plain without textures or other natural scenes commonly found in real-world documents. Furthermore, we did not apply image augmentations, such as skewing, rotation, or perspective distortion, which could further challenge model robustness.

Looking ahead, we are extending this work in two key directions. First, we are currently developing a Pashto VQA dataset and benchmark. Second, we are working on the first large-scale handwritten Pashto OCR dataset. Furthermore, we plan to develop a more realistic version of the PsOCR dataset by adding different background patterns and diverse lighting conditions to the images, to more accurately mimic real-world conditions. Together, these efforts will deepen our understanding of multimodal model performance on low-resource languages and drive future improvements in Pashto document analysis.

\section{Data Availability}
The PsOCR dataset is divided into two parts: 1) the PsOCR Benchmark, comprising 10K images for evaluation, which is publicly available on HuggingFace\footnote{\underline{https://huggingface.co/datasets/zirak-ai/PashtoOCR}} and Kaggle\footnote{\underline{https://www.kaggle.com/datasets/drijaz/PashtoOCR}}; and 2) the PsOCR Train Set, containing approximately one million images, which is available upon request by emailing the corresponding author.

\end{document}